\pdfoutput=1

\documentclass[11pt]{article}

\usepackage{acl}
\usepackage{amsmath}
\usepackage{amsfonts}

\usepackage{times}
\usepackage{latexsym}

\usepackage[T1]{fontenc}

\usepackage[utf8]{inputenc}

\usepackage{microtype}

\usepackage{inconsolata}

\usepackage{graphicx}

\title{KoGNER: A Novel Framework for Knowledge Graph Distillation on Biomedical Named Entity Recognition}

\author{
    \textbf{Heming Zhang\textsuperscript{1}*}, 
    \textbf{Wenyu Li\textsuperscript{2,3}*}, 
    \textbf{Di Huang\textsuperscript{1,2}*}, 
\\  
    \textbf{Yinjie Tang\textsuperscript{3}}, 
    \textbf{Yixin Chen\textsuperscript{2}}, 
    \textbf{Philip Payne\textsuperscript{1}},
    \textbf{Fuhai Li\textsuperscript{1}†}
\\
    \textsuperscript{1}Institute for Informatics, Data Science and Biostatistics, Washington University in St. Louis,
\\
    \textsuperscript{2}Department of Computer Science and Engineering, Washington University in St. Louis, 
\\
    \textsuperscript{3}Energy, Environmental and Chemical Engineering, Washington University in St. Louis
\\
    \small{
       \textbf{†Correspondence:} \href{mailto:fuhai.li@wustl.edu}{fuhai.li@wustl.edu}
    }
\\
    \small{
        \textsuperscript{*}Equal contribution as co-first authors.
    }
}

\begin{document}
\maketitle
\begin{abstract}
Named Entity Recognition (NER) is a fundamental task in Natural Language Processing (NLP) that plays a crucial role in information extraction, question answering, and knowledge-based systems. Traditional deep learning-based NER models often struggle with domain-specific generalization and suffer from data sparsity issues. In this work, we introduce \textbf{K}n\textbf{o}wledge \textbf{G}raph distilled for \textbf{N}amed \textbf{E}ntity \textbf{R}ecognition (KoGNER), a novel approach that integrates Knowledge Graph (KG) distillation into NER models to enhance entity recognition performance. Our framework leverages structured knowledge representations from KGs to enrich contextual embeddings, thereby improving entity classification and reducing ambiguity in entity detection. KoGNER employs a two-step process: (1) Knowledge Distillation, where external knowledge sources are distilled into a lightweight representation for seamless integration with NER models, and (2) Entity-Aware Augmentation, which integrates contextual embeddings that have been enriched with knowledge graph information directly into GNN, thereby improving the model's ability to understand and represent entity relationships. Experimental results on benchmark datasets demonstrate that KoGNER achieves state-of-the-art performance, outperforming finetuned NER models and LLMs by a significant margin. These findings suggest that leveraging knowledge graphs as auxiliary information can significantly improve NER accuracy, making KoGNER a promising direction for future research in knowledge-aware NLP.
\end{abstract}

\begin{figure*}[t]
    \centering
    \includegraphics[width=\textwidth]{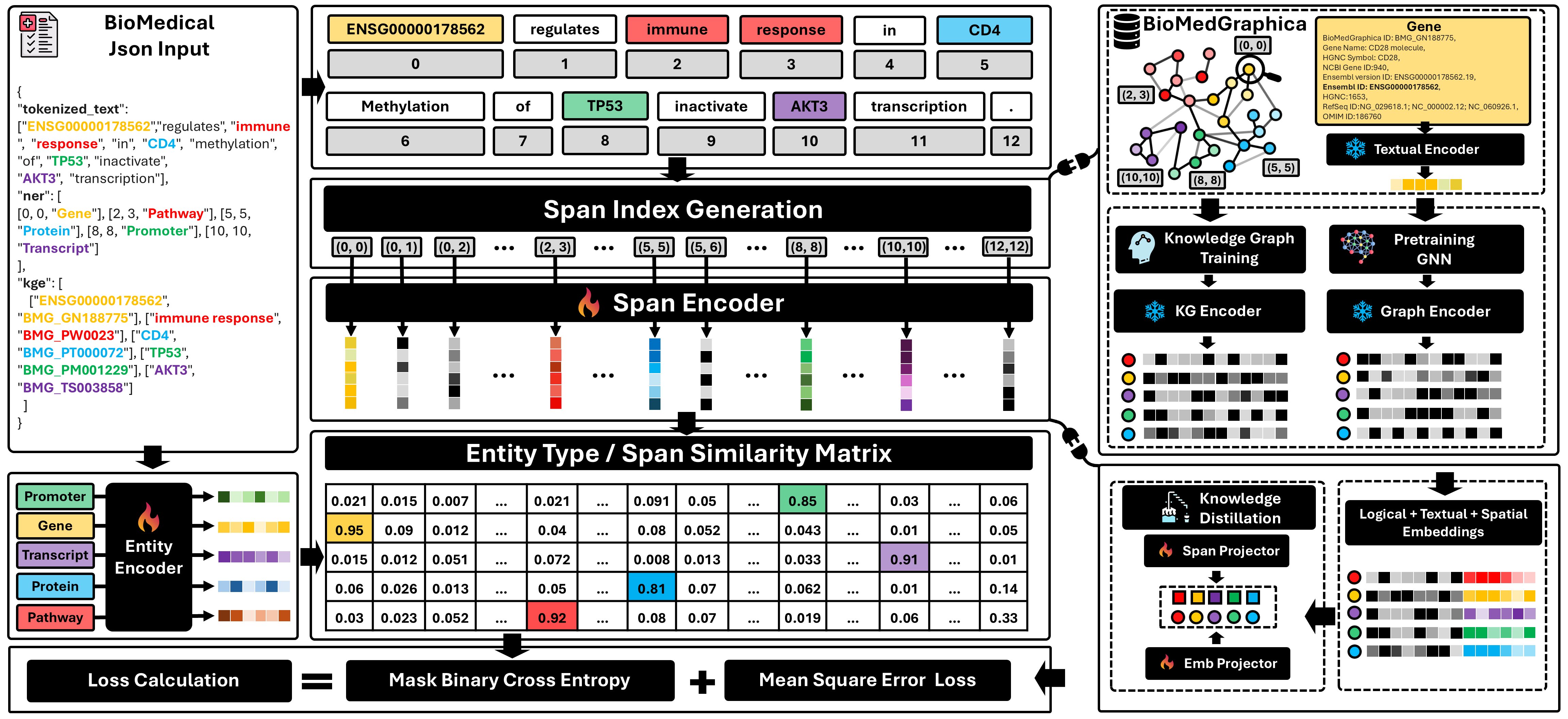} 
    \caption{Overall architecture of KoGNER}
    \label{fig:yourlabel}
\end{figure*}

\section{Introduction}

The rapid evolution of large language models (LLMs) has led to a paradigm shift in information extraction \cite{wang2023gptnernamedentityrecognition}. With models like GPT-4o \cite{openai2024gpt4ocard} and emerging open-source alternatives like LLaMA 3.3 \cite{grattafiori2024llama3herdmodels}, DeepSeek v3 \cite{deepseekai2024deepseekv3technicalreport}, etc., it is now possible to extract an unlimited variety of entity types using natural language instructions - a capability that traditional NER systems, limited to a fixed set of predefined entities, cannot match \cite{kim-etal-2024-verifiner}, except for GLiNER \cite{zaratiana2023gliner} in a zero-shot setting. However, the promise of powerful LLM models is tempered by practical challenges: Their enormous parameter counts necessitate vast computational resources, which in turn escalate costs when deployed at scale. Conventional NER methods, including rule-based systems and machine learning models, deliver high accuracy, clear interpretability, and efficient computational performance \cite{zhou2022conner}, especially in practical application scenario \cite{10.1145/3459637.3482491}. Yet expanding their scope traditionally demands labor-intensive data annotation and retraining. Recent advances have sought to reconcile these divergent approaches. Innovations such as SCANNER \cite{ok-etal-2024-scanner}, KCL \cite{zhang-etal-2024-kcl}, UniNER \cite{zhou2024universalner}, and KAWR \cite{he2020knowledge} demonstrate that integrating structured external knowledge with the adaptive capabilities of NER systems can substantially enhance the performance and flexibility of entity and relation extraction. In addition, entity and relation extraction also provides the basis for some high-level applications such as Knowledge Graph
construction \cite{burns2024discoveringlatentknowledgelanguage}.

Knowledge distillation has emerged as a powerful technique for bridge-building the gap between high-performance, large-scale models and the practical demands of resource-constrained applications. This approach enables the student model to approximate the performance of the teacher while significantly reducing computational and memory overhead. Existing research \cite{zhou-etal-2021-multi} has shown that integrating CRF cross-entropy with fuzzy learning effectively facilitates the transfer of knowledge from large pre-trained models to lightweight NER systems. SKD-NER \cite{chen-he-2023-skd} also integrates reinforcement learning into the knowledge distillation process within a span-based continual learning framework. However, they lack domain-specific knowledge and cannot generalize to new entity types in a zero-shot fashion.

Within the biomedical domain, the intricate structure of electronic patient records and clinical data, the nested organization of named entities, and the abundance of specialized terminology poses distinct challenges for NER task. The intricate biological mechanism also poses difficulties for entity relationship extraction. Recent advances in the use of LLM both directly \cite{doi:10.1021/acssynbio.3c00310} and indirectly \cite{https://doi.org/10.1002/advs.202306724} to extract biomedical information have shown significant promise in overcoming these challenges by providing structured information curated by humans or finetuning with domain specific dataset. Here, we offer an approach that uses the best sides of both human-curated knowledge graphs and language model's textual representation through a knowledge distillation method. This approach synergizes traditional NER techniques with the dynamic extraction capabilities of knowledge graph distillation. To the best of our knowledge, this is the first work to employ both knowledge graphs and language models to construct a comprehensive database for biomedical NER. Our framework integrates text-attributed knowledge graphs specifically tailored to address the unique challenges of biomedical text. By reimagining the role of knowledge graphs in NER, we advance the state-of-the-art in biomedical domain-specific information extraction.

\section{Method}
This section introduces KoGNER, a model designed to enhance biomedical named entity recognition (NER) by leveraging distilled knowledge, spatial, and logical information from a large biomedical knowledge graph. By embedding these structured insights into a bi-encoder framework\cite{zaratiana-etal-2024-gliner}, KoGNER ensures rich, context-aware entity representations, improving both accuracy and generalizability across diverse biomedical domains.

\subsection{Texutal BiEncoder}
Given text input $\mathcal{X}=[x_1, x_2, \cdots, x_t,\cdots, x_T]$ and corresponding entity set $\mathcal{N}=[n_1, n_2,\cdots,n_c,\cdots, n_C]$,  where $T$ is the number of tokeinized words and $C$ is the count of entities. The span indices are produced by establishing a search window of a predetermined width $w$, with $\mathcal{I}=\{(0,0),(0,1),\cdots,(t,t+1),\cdots, (T-1,T+w-1)\}=\{i_{pq}\}_{0}^{T\times{w}}$. Afterwards, the input text $\mathcal{X}$ and entities $\mathcal{N}$ are embedded into their latent representations $\mathcal{X'}=\{{x'}_p\}_{0}^{T}\in{\mathbb{R}^{T\times{d}}}$ and $\mathcal{N'}=\{n'_{p}\}^{C}_{0}\in{\mathbb{R}^{|\mathcal{B}|\times{d}}}$, where $\mathcal{B}$ denotes the union of type categories associated with the entities. The encoded input text data, $\mathcal{X'}$ is then combined with span index set $\mathcal{I}$ to yield a span representation $\mathcal{S}=\{s_{pq}\}_{0}^{T\times{w}}\in{\mathbb{R}^{(T\times{w})\times{d}}}$ computed via $s_{pq}=\text{FFN}(x'_{p}\otimes{x'_{q}})$ with $\text{FFN}$ indicating a two-layer feedforward neural network and $\otimes$ representing the concatenation. Concurrently, the entity representations $\mathcal{N'}$ will also be encoded via the two-layer FFN.

\subsection{Language Loss} 
To evaluate whether a span \((p, q)\) corresponds to entity type \(c\), we calculate the following matching score:

\begin{equation}
\phi(p, q, b) = \sigma(s^{T}_{pq}{n'}_{b}) \in \mathbb{R}
\end{equation}

where \(\sigma\) is the sigmoid activation function. Since we use binary cross-entropy loss during training, \(\phi(p, q, b)\) represents the probability that the span \((p, q)\) is of type \(b\). Thus, the language loss $\mathcal{L}_{\text{lang}}$ comprising span index set $\mathcal{I}$ and entity types $\mathcal{B}$, is defined as:

\begin{align}
\mathcal{L}_{\text{lang}} = & - \sum_{a \in {\mathcal{I}} \times {\mathcal{B}}} \Big( 
\mathbb{I}_{a \in {\mathcal{P}}} \log \phi(a)  \notag \\ 
& \quad + \mathbb{I}_{a \in \mathcal{N}} \log (1 - \phi(a)) \Big)
\end{align}

where the variable \(a\) represents a pair of span/entity type, such as $(p, q, c)$, and $\mathbb{I}$ is an indicator function, which returns 1 when the specified condition is true and 0 otherwise for belonging to positive samples. This loss function corresponds to binary cross-entropy.

\subsection{Distillation Loss} 
During knowledge distillation, we designate the teacher model as the embedding derived from the pretrained graph neural network (GNN) and knowledge graph encoder models. Together with input tex , we add knowledge entities input as $\mathcal{V}_{sub}=\{(i_1, v^{sub}_1),\cdots,(i_c, v^{sub}_c),\cdots,(i_C, v^{sub}_C)\}$, where $i_c$ correpsonds certain span index $i_{pq}$, and $v_c$ denotes the corresponding entity in the knowledge graph, $\mathcal{G}=\{\mathcal{V,E}\}$, where $|\mathcal{V}|=M$ and $|\mathcal{E}|=e$. For each entity in the knowledge graph, it has the corresponding name descriptions, denoted as $z$ and correpsonding entity types, denoted as $y$. For those named description, it will be encoded by combining those name descriptions for each entity with $z'\in{\mathbb{R}^{d'}}$. And all of the encoded name descritptions for those entities will be comprised as the node features $Z=\in{\mathbb{R}^{M\times{d'}}}$, and all of the entity types will be collected as the node / vertex label set $Y\times{\mathbb{R}^{M}}$. Therefore, we can pretrain a graph neural network (GNN) by

\begin{equation}
    \hat{Y} = \text{GNN}(Z,\mathcal{E};K)
\end{equation}

where $\text{GNN}$ is the model need to be pretrained, which is the node classification model and $K$ is the number of layer for the message propagation. And the pretrained model will be marked as $\text{GNN}_{\text{pre}}$. And this pretraining process will ensure the model can capture the spatial information from the knowledge graph. Then, the message propagation will be made by

\begin{equation}
    Z'=\text{GNN}_{\text{pre}}(Z,\mathcal{E};K)
\end{equation}

, where $Z'\in{\mathbb{R}^{M\times{d'}}}$. Additionally, the knowledge graph inherently captures the logical relationships between entities, $\mathcal{V}$ in the graph. Herein, the knowledge graph can be trained via \textbf{TransR}\cite{zaratiana2023gliner}, which can ensure model can capture symmetrical and 1-to-N relations in the knowledge graph $\mathcal{G}$. Therefore, the logical embeddings for entities will be generated by 

\begin{equation}
    Z''=\text{TransR}(\mathcal{V,E})
\end{equation}

, where $Z''\in{\mathbb{R}^{M\times{d''}}}$. Then, the node embeddings will be concatenate by $H=[Z,Z',Z'']\in{\mathbb{R}^{M\times{r}}}$, $r=2d'+d''$, which means this embedding contain the textual, spatial and logical information for all entities. 

By extracting the represented span embeddings, ${s_{pq}}$ from certain identifier $i_c$ (corresponding to $i_{pq}$), each of those extracted span embeddings will be combined as $\mathcal{S}_{sub}=\{s_{pq}\}_0^{C}\in{\mathbb{R}^{C\times{d}}}$. Meanwhile, the corresponding embeddings from the knowledge graph will be extracted by $\mathcal{V}_{sub}$ with $H_{sub}\in{\mathbb{R}^{C\times{r}}}$. Then, the distillation will be made to make sure those information will be distilled into the KoGNER by

\begin{align}
    \mathcal{L}_{\text{dist}} = 
    & \ \text{MSE} \Big( \text{MLP}_{\text{span}}(\mathcal{S}_{sub}),  \notag \\ 
    & \quad \text{MLP}_{\text{dist}}(H_{sub}) \Big)
\end{align}

where MSE is the mean square error loss and MLP denotes the linear transformation, which will map entity embeddings from knowledge graph and span representation into the same dimensions. In the end, the overall loss function will be

\begin{equation}
    \mathcal{L}=\mathcal{L}_{\text{lang}} + \mathcal{L}_{\text{dist}}
\end{equation}

\section{Experimental Settings and Results}
In this experiment, we utilized a teacher-student framework for knowledge distillation. The teacher model comprises a Graph Transformer \cite{shi2020masked} for GNN pretraining and TransR for the knowledge graph encoder, ensuring effective relational modeling (refer to the Appendix for details on teacher model pretraining). For the student model, we selected GliNER due to its strong generalization capability, as evidenced by its high zero-shot performance across multiple NER datasets. The model configuration and hyperparameter settings are provided in the Appendix.

To facilitate distillation training, we generated the BMG dataset (see Appendix) and used the F1 score as the evaluation metric. We conducted a zero-shot evaluation on the BMG dataset, testing both mainstream closed-source LLMs, including the OpenAI GPT family and Claude, as well as open-source LLMs such as LLaMA and the NER model GliNER. As shown in \hyperlink{tab:med-gpt}{Table 1}, GPT-4o achieves the highest F1-score of 10.3\%. However, despite this result, the findings suggest that these models are not suitable for out-of-the-box usage and highlight the complexity of the BMG dataset, which requires careful fine-tuning for optimal performance.

\begin{table}[ht]
  \centering
  \hypertarget{tab:med-gpt}{
  \begin{tabular}{lc}
    \hline
    \textbf{Model} & \textbf{BMG}   \\
    \hline
    GPT-4o    & {10.3}          \\
    GPT-4o-mini & {8.8}                \\
    Claude3.5-sonnet    & {8.8}            \\
    GLiNER  & {6.0}         \\
    LLaMA3.3-70b      & {5.4}          \\
    \hline
  \end{tabular}
  }
  \caption{Results of comparing relavent models in zero-shot NER task on BMG dataset. Prompts used for testing LLMs follows \cite{zhou2024universalner}.}
  \label{tab:med-gpt}
\end{table}

\begin{table}[ht]
  \centering
  \hypertarget{tab:med-baselines}{%
  \begin{tabular}{lcc}
    \hline
    \textbf{Dataset} & \textbf{KoGNER} & \textbf{GLiNER (ft)}  \\
    \hline
    BMG      & {\textbf{51.6}} & {42.8}     \\
    GENIA    & {\textbf{32.8}} & {31.1}     \\
    NCBI     & {32.6}         & {\textbf{44.1}}     \\
    BC5CDR   & {43.2}         & {\textbf{55.8}}     \\
    BC4CHEMD & {33.0}         & {\textbf{39.3}}     \\
    BC2GM    & {\textbf{31.2}} & {29.8}     \\
    \hline
  \end{tabular}%
  }
  \caption{Results of testing 6 datasets on KoGNER and a BMG-finetuned version of GLiNER.}
  \label{tab:med-baselines}
\end{table}

Compared with a BMG-finetuned version of GLiNER, GLiNER (ft), on six diverse datasets, KoGNER continues to demonstrate robust performance on BMG, GENIA and BC2GM. However, in the NCBI, BC5CDR, and BC4CHEMD datasets, the fine-tuned GLiNER achieves higher scores. These findings suggest that while KoGNER's zero-shot capabilities are effective in scenarios with a broad range of entity type labels, fine-tuning with domain-specific data can provide notable advantages for more complex tasks, particularly in disease and drug recognition. \hyperlink{tab:med-baselines}{Table 2} also indicates that KoGNER demonstrates strong performance on datasets such as BMG, GENIA, and BC2GM. However, on NCBI, BC5CDR, and BC4CHEMD, GLiNER (ft) achieves higher scores. This suggests that while KoGNER is robust and effective in some scenarios with more entities type labels, fine-tuning with domain-specific data may offer advantages in prediction for diseases and drugs.

\section{Conclusion}
Our study presents KoGNER, a novel distillation framework for Named Entity Recognition (NER) models, specifically designed to enhance biomedical entity recognition. By integrating KoGNER with the BioMedGraphica (BMG) dataset, we successfully leverage structured knowledge from knowledge graphs to improve entity detection and classification.

The experimental results confirm that KoGNER achieves superior performance compared to traditional and large language model (LLM)-based NER approaches. Our approach enhances entity embeddings with structured, knowledge-rich information, allowing the model to better contextualize and disambiguate biomedical entities. This is particularly crucial for biomedical text processing, where complex, nested, and highly specialized terminologies often pose challenges for standard NER models.

Moreover, our knowledge distillation strategy ensures scalability and efficiency, enabling KoGNER to operate with reduced computational overhead while maintaining state-of-the-art accuracy. The results indicate that our framework is highly effective in zero-shot NER tasks, demonstrating strong generalization across diverse biomedical datasets. These findings suggest that knowledge graph integration is a promising direction for enhancing domain-specific NER tasks and could be extended to other structured data-driven NLP applications.

\section{Ethics Statement}
Our research was conducted in full adherence to rigorous ethical standards. We utilized publicly available, licensed biomedical datasets and ensured that all data were anonymized and devoid of any personally identifiable information. In cases where the original data collection involved human subjects, appropriate informed consent was obtained by the data providers. Throughout the development of BioMedGraphica, we maintained transparency in our methodologies and prioritized data integrity and privacy. Lastly, the GLiNER framework was developed under the Apache License Version 2.0, January 2004, ensuring open collaboration and adherence to industry best practices.

\section*{Limitations}
Despite its strengths, KoGNER faces several challenges. The complexity of the BMG dataset makes entity recognition difficult, often leading to truncated or misclassified labels in large language models (LLMs). Models like GPT-4o and Claude 3.5 underperform in zero-shot biomedical NER, highlighting the need for domain-specific fine-tuning. Additionally, the presence of nested entities in biomedical text poses a challenge, as standard NER models struggle with overlapping entity spans, suggesting that testing nested-aware architectures could enhance accuracy. While KoGNER excels in zero-shot settings, fine-tuned models such as GLiNER (ft) perform better in specific domains like disease and drug recognition, emphasizing the importance of task-specific optimization. Furthermore, the integration of graph neural networks (GNNs), while beneficial for knowledge augmentation, introduces computational overhead, which could be mitigated through more efficient knowledge distillation techniques. Future research should explore nested NER models, improved fine-tuning strategies, and optimized distillation methods to further refine KoGNER’s performance in biomedical applications.

\section*{Acknowledgments}
This research was partially supported by NIA R56AG065352, NIA 1R21AG078799-01A1, NINDS 1RM1NS132962-01, and NLM 1R01LM013902-01A1.

\bibliography{custom}

\appendix
\section{Example Appendix}
\label{sec:appendix}
During knowledge distillation, we designate the teacher model as the embedding derived from the knowledge graph embedding model, TransR, while the student model is the base NER model, GLiNER. We employ feature-based distillation by computing the mean squared error (MSE) loss between the knowledge graph embeddings and the NER model embeddings.

We utilized the following parameters to distill knowledge from a biomedical knowledge graph into an NER model to obtain KoGNER.

\section{Data Preparation}

\begin{table}
  \centering
  \begin{tabular}{lc}
    \hline
    \textbf{Parameter} & \textbf{Value} \\
    \hline
    Base Embed & \verb|deberta-v3-large| \\
    Max Span Width & 8 \\
    Hidden Size & 512 \\
    KG Hidden Size & 58 \\
    Dropout & 0.4 \\
    \hline
    \# Steps & 10000 \\
    Train Batch & 8 \\
    Warm Up Ratio & 0.1 \\
    Learning Rate & Cosine \\
    \hline
    Loss a\_cls & -1 \\
    Loss a\_BCE & 0.8 \\
    Loss b\_dist & 0.2 \\
    \hline
    Gliner Path & \verb|gliner_large-v2.5| \\
    \hline
  \end{tabular}
  \caption{Model hyperparameters and configurations.}
  \label{tab:model_params}
\end{table}
\subsection{BMG NER Training Data Processing}
Data preprocessing for the BioMedGraphica \cite{Zhang2024.12.05.627020} database involved a multi-step pipeline designed to handle the extensive and heterogeneous biomedical relationship data efficiently. BMG dataset consists of For each relationship edge type, a random subsample—approximately one in every 10,000 rows was extracted to create a representative dataset.  Next, a mapping function was employed to dynamically select the correct column names for the entity types based on the their relationship type, ensuring that the appropriate fields were used for entity extraction. For each sampled row, the first non-null entity from each corresponding CSV file was retrieved by splitting and filtering the cell contents. Finally, these extracted entity pairs, along with their associated relationship types and unique identifiers, were aggregated and fed into the LLM data generation pipeline.
\subsection{LLM-assisted NER Data Generation}
\begin{figure}[t]
  \includegraphics[width=\columnwidth]{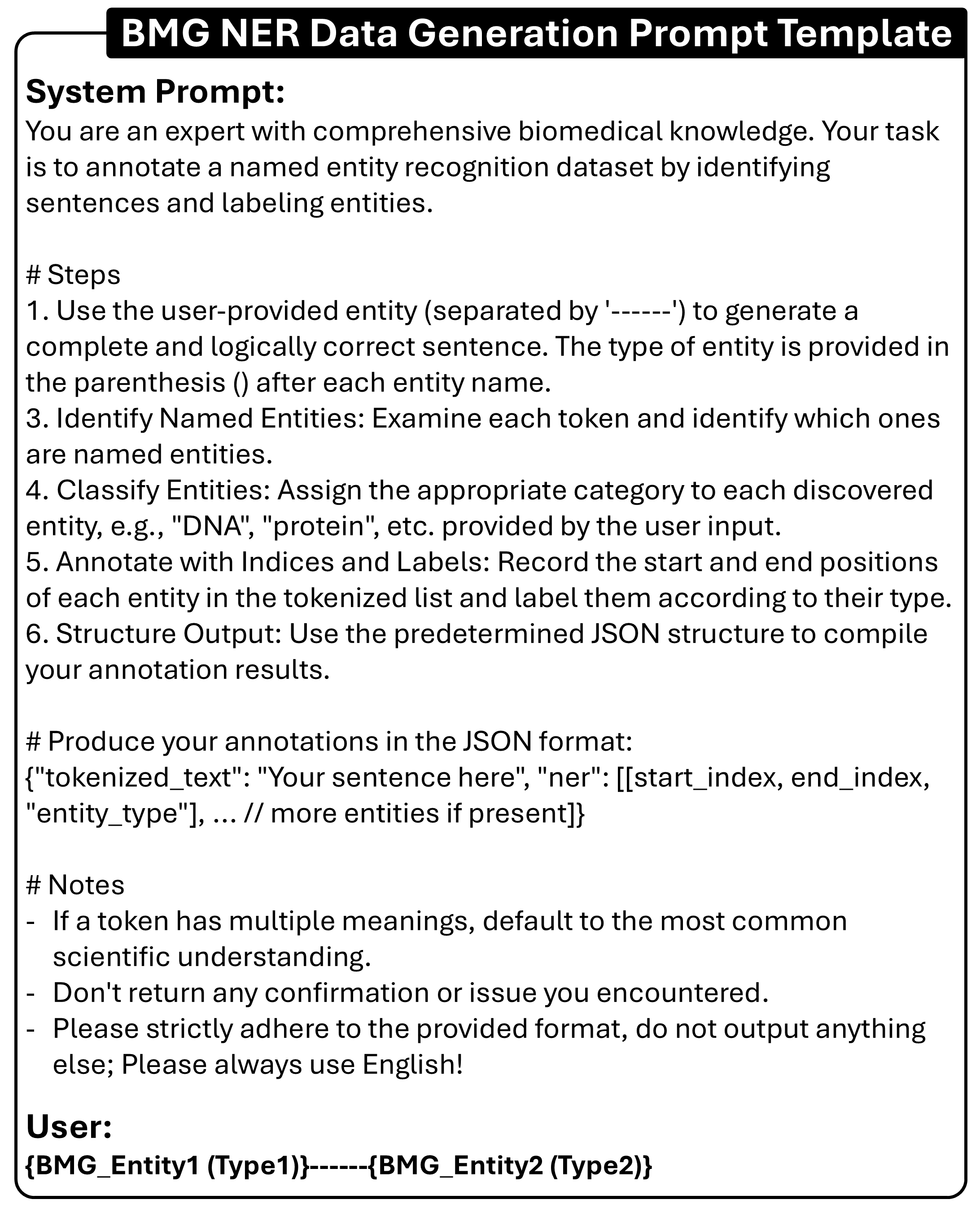}
  \caption{Step-by-step data construction prompt for LLM to generate synthetic sentences consisting of BMG entities filtered from public databases, span indices, and the corresponding type in JSON format. The Knowledge Graph Embedding ("kge") tag is added according to BMG dataset.}
  \label{fig:experiments}
\end{figure}
This section implements a pipeline that leverages an LLM to generate NER sentences enriched with knowledge graph information based on input entity data. First, it uses a pre-defined system prompt (Figure 2) and iterates over the preprocessed dataset obtaiend from section B.1. containing entity relationships. For each row, it constructs a user prompt by formatting entities after cleaning extraneous characters and appending their corresponding relationship types. For each answer, augments it by adding a "kge" field that contains the entity names and their IDs from BMG.

A sanity check takes a dictionary (entry) that is expected to have the same format defined in Figure 2. To ensure that the entity texts (from both the NER and KGE annotations) can be accurately located within the tokenized text, and that their spans are correctly identified, the check is implemented as follows:
\begin{itemize}
    \item \textbf{Extracts Textual Representation}: Retrieves the textual representation of the first and second entities from the \texttt{kge} field.
    \item \textbf{Tokenization}: Tokenizes the entry’s \texttt{tokenized\_text} using \texttt{tokenize\_text} to ensure consistent tokenization and also tokenizes individual entity strings.
    \item \textbf{Finding Entity Matches}: Iterates over the tokenized text to locate the first occurrence where the sequence of tokens matches the entity tokens.
    \item \textbf{Appending Entity Spans}: When a match is found, appends a span (start and end token indices) with the corresponding entity type from the \texttt{ner} field to the \texttt{entity\_spans} list and a span with an identifier from the \texttt{kge} field to the \texttt{kges} list.
    \item \textbf{Handling Index Errors}: If an \texttt{IndexError} occurs due to missing or unexpected structures in \texttt{ner} or \texttt{kge}, prints debugging information including the full text, search substring, and found index before continuing.
    \item \textbf{Ignoring Type Errors}: Silently ignores any \texttt{TypeError}s encountered during extraction, such as missing keys or an unexpected structure.
\end{itemize}

\end{document}